\documentclass{article}

\usepackage[nonatbib, preprint]{neurips_2024}

\usepackage[utf8]{inputenc} 
\usepackage[T1]{fontenc}    
\usepackage[allcolors=blue,colorlinks=true]{hyperref}       
    
\usepackage{url}            
\usepackage{booktabs}       
\usepackage{amsmath}       
\usepackage{amsfonts}       
\usepackage{nicefrac}       
\usepackage{microtype}      
\usepackage{xcolor}         
\usepackage{graphicx}

\usepackage[frozencache=true,cachedir=.]{minted}

\usepackage[sorting=none, style=numeric-comp]{biblatex}
\title{MESS: Modern Electronic Structure Simulations}

\author{%
  Hatem Helal\\
  Graphcore\\
  \texttt{hatemh@graphcore.ai}\\
\And
  Andrew Fitzgibbon\\
  Graphcore\\
  \texttt{awf@graphcore.ai}\\
}

\begin{document}

\maketitle

\newminted{python}{style=borland,xleftmargin=16pt}
\newmintinline{python}{}

\begin{abstract}
  Electronic structure simulation (ESS) has been used for decades to provide quantitative scientific insights on an atomistic scale, enabling advances in chemistry, biology, and materials science, among other disciplines.
  Following standard practice in scientific computing, the software packages driving these studies have been implemented in compiled languages such as FORTRAN and C.  However, the recent introduction of machine learning (ML) into these domains has meant that ML models must be coded in these languages, or that complex software bridges have to be built between ML models in Python and these large compiled software systems.
  This is in contrast with recent progress in modern ML frameworks which aim to optimise both ease of use and high performance by harnessing hardware acceleration of tensor programs defined in Python.
  We introduce MESS: a modern electronic structure simulation package implemented in JAX; porting the ESS code to the ML world. 
  We outline the costs and benefits of following the software development practices used 
  in ML for this important scientific workload.
  MESS shows significant speedups on widely available hardware accelerators and simultaneously opens a clear pathway towards combining ESS with ML.
  MESS is available at \url{https://github.com/graphcore-research/mess}.
\end{abstract}

\section{Introduction}
Approximating the quantum-mechanical behaviour of electrons in atoms, molecules, and solids 
is a long standing central challenge in the physical sciences \cite{dirac1929quantum}.  
Overcoming this challenge would provide a deeper understanding of the 
atomistic processes that underpin diverse phenomena.
For example, these simulations could be used to efficiently screen
drug candidates through direct modelling of their interactions with 
biological targets at the molecular level \cite{batra2020screening}.
In materials science, this understanding could accelerate the design of advanced materials for renewable energy by precisely tailoring their electronic properties \cite{gomez2016design, shipley2021high}.
Density Functional Theory (DFT) has emerged as a leading approach for ESS,
striking a favourable balance between computational cost and accuracy when compared to more 
computationally demanding methods like coupled cluster theory or configuration interaction 
\cite{keith2021combining, szabo2012modern}.

A recent high-impact research direction in computational chemistry has been the development of simulation environments that integrate ML models \cite{keith2021combining, von2020retrospective, westermayr2021perspective}. 
While significant progress has been made by establishing complex software bridges between traditional ESS
codes and ML frameworks, this approach presents a steep learning curve for researchers.
These scientists must become proficient in not only
electronic structure theory and its application to atomistic systems, 
but also the intricacies of ML model training and deployment, 
as well as high-performance scientific computing techniques. 
This multifaceted expertise requirement acts as a significant barrier, 
hindering the broader adoption and potential impact of these powerful simulation tools.

To address these challenges and accelerate this naturally interdisciplinary research, 
we introduce MESS: a Modern Electronic Structure Simulation framework. 
MESS is designed to bridge the gap between electronic structure methods and machine learning models by enabling 
their integration within a single unified environment.

We summarise our contributions as follows:
\begin{itemize}
    \item Implementation of \textbf{end-to-end differentiable} electronic structure simulations, enabling algorithmic gradient evaluation for routine tasks such as electronic energy minimisation and atomic force evaluation.
    \item Development of a \textbf{high-throughput DFT simulation} framework leveraging hardware acceleration and batch processing for efficient exploration of configurational space.
    \item Creation of an \textbf{extensible and modular simulation environment} that facilitates the seamless integration of machine learning models into electronic structure calculations.
\end{itemize}
MESS is released as an open-source Python package under the permissive MIT license, 
with all examples and benchmarks described here provided 
as executable notebooks for reproducible results and interactive exploration.
\footnote{Notebooks are available as documentation pages at \url{https://graphcore-research.github.io/mess}}
Additionally, by leveraging the powerful libraries of the JAX ecosystem, MESS 
demonstrates the potential of adapting software abstractions developed for
machine learning to enhance and accelerate computational chemistry workloads.

\section{Related Work}

DFT in principle provides an exact solution to the electronic structure problem by placing the electron density on par with solving the Schr\"odinger equation for the 
many-body wavefunction \cite{hohenberg1964inhomogeneous, kohn1965self}.
This dramatic simplification is possible by the introduction of the exchange-correlation functional
that accounts for the quantum-mechanical nature of electron-electron interactions.
However, the exact form of the universal exchange-correlation functional is still unknown \cite{burke2013dft}.
Practical implementations of DFT require selecting an approximate exchange-correlation functional, 
introducing a critical trade-off between accuracy and computational cost \cite{rappoport2008functional}.  

With hundreds of such functionals available, navigating the vast landscape of options and selecting the most appropriate form for a specific application can become a laborious task even for experts in the field 
\cite{lehtola2018recent, rappoport2008functional}.
Numerous studies have attempted to benchmark and classify these functionals, offering guidance on their strengths and weaknesses for different types of systems and properties 
\cite{pribram2015dft, peverati2021fitting, goerigk2019trip, kim2013understanding, kalita2022well}. 
Recent efforts have leveraged machine learning and offer a promising route towards a computationally efficient
approximation to the universal functional \cite{kirkpatrick2021pushing, li2021kohn}.

While DFT simulations offer valuable insights into the electronic structure and static properties of molecules,
they do not inherently capture the dynamic evolution of these systems over time.  
To study the time-dependent behavior of atomistic structures, such as the breaking and forming of chemical bonds, molecular dynamics (MD) simulations are employed \cite{eastman2017openmm, thompson2022lammps, case2023ambertools}. 
In MD simulations, the forces acting on atoms can be derived from a variety of models, ranging from simple analytic functions to more sophisticated quantum-mechanical calculations \cite{car2006ab}. 
These forces are then used to numerically integrate the equations of motion, allowing the trajectories of atoms to be followed over time. 
This approach provides a powerful tool for investigating a wide range of phenomena, from chemical reactions to the self-assembly of complex materials \cite{lindorff2011fast}.

DFT-based MD simulations offer high accuracy in capturing the intricate electronic interactions that govern chemical bonding and reactivity.
However, their widespread adoption has been hampered by the inherent computational cost of repeatedly solving the electronic structure problem at each time step. 
This limitation has typically restricted DFT-based MD simulations to relatively small systems and short timescales \cite{musaelian2023scaling}.
To overcome this bottleneck and enable simulations of large-scale systems over extended periods, researchers have
utilised computationally cheaper alternatives, such as hand-crafted interatomic potentials
and targeting specialised hardware architectures 
\cite{klepeis2009long, shaw2010atomic, shaw2014anton}.

One promising solution that has emerged in recent years are 
Machine Learning Interatomic Potentials (MLIPs)
\cite{behler2007, bartok2010gaussian, thompson2015spectral, schutt2018schnet, deringer2019machine, drautz2019atomic, von2020retrospective, batzner20223}.
These potentials leverage the flexibility of machine learning algorithms to learn 
complex relationships between atomic configurations and their corresponding energies and forces, 
as calculated by DFT. 
By utilising supervised learning on vast datasets of DFT simulations, 
MLIPs can accurately reproduce the potential energy surface of a system, enabling efficient and accurate MD simulations 
\cite{eastman2023spice, batatia2023foundation}. 
MLIP-driven MD simulations offer a substantial advantage over pure DFT-based MD,
enabling the study of significantly larger systems and longer timescales 
due to their reduced computational demands \cite{musaelian2023scaling, jia2020pushing}.

Various permutations of the combined features of hardware acceleration, automatic differentiation, and
dynamic interpreted programs have been previously explored for ESS.  
TeraChem pioneered ESS on GPU accelerators
by implementing these calculations directly in the CUDA framework 
\cite{seritan2021terachem, nickolls2008scalable}.
Being a closed-source project inherently limits the scope of algorithmic research that can be 
explored with TeraChem but it has been applied to investigate quantum simulations of proteins
\cite{kulik2012ab}.
PySCF supports modular ESS workflows through the familiar and easy-to-use NumPy interface \cite{sun2018pyscf}.
Extensions to PySCF have been developed to support automatic differentiation and GPU acceleration  \cite{zhang2022differentiable, wu2024python}.
Automatic differentiation is one of the pillars of modern machine learning 
and has recently been explored for ESS \cite{baydin2018automatic, tan2023automatic}.  JAX MD implements a differentiable simulation environment with straightforward interoperability between MLIPs and MD time-integrators \cite{schoenholz2020jax}.  
Differentiable implementations for the Hartree-Fock method have 
been explored \cite{tamayo2018automatic, yoshikawa2022automatic}.
These projects demonstrated the applicability of automatic differentiation for solving this particular instantiation of electronic
structure simulation.
The DQC project demonstrated that a differentiable DFT implementation
enables learning the exchange-correlation functional from data \cite{kasim2022dqc}.  
DQC uses the libcint library to evaluate the molecular integrals which 
prevents utilising modern hardware accelerators \cite{sun2015libcint}.

To the best of our knowledge, MESS is the first project to fully 
integrate end-to-end differentiable high-level programs
with hardware-agnostic acceleration through compilation.

\section{Electronic Structure Theory in a Nutshell}

The central problem of electronic structure is solving the Schr\"odinger equation for
the ground-state configuration of electrons given a molecular structure of interest.  
We will focus the discussion here on simulations within the Born-Oppenheimer approximation \cite{martin2020electronic}.
The electron representation in MESS is flexible and the examples shared here 
use localised molecular orbitals, represented as a linear combination of atomic orbitals (LCAO) \cite{szabo2012modern}
\begin{equation}
    \psi_i(\mathbf{{r}}) = \sum_\mu C_{\mu i} \phi_\mu(\mathbf{r} - \mathbf{R}_{\mu}; Z_{\mu}),
\end{equation}
where the $\{\psi_i\}$ are the set of orthonormal molecular orbitals that are expressed
as an expansion in terms of a basis set of atomic orbitals $\{\phi_\mu\}$.
Each atomic orbital $\phi_\mu$ is associated with a unique atom which
defines the centre $\mathbf{R}_\mu$ and the atomic number $Z_\mu$.
We have some freedom in how we select the basis set of atomic orbitals and, following standard practice, MESS uses Gaussian orbitals
\cite{szabo2012modern}.
The specific parametrisation of the basis set is a critical decision that significantly 
impacts the accuracy and computational cost of the simulation.
A plethora of pre-computed basis sets are available, 
ranging from minimal sets designed for rapid calculations to larger, 
more comprehensive sets that offer increased accuracy at the expense of computational resources.
MESS interfaces with the basis set exchange which provides the pre-computed parameters defining the atomic orbitals \cite{pritchard2019new}.

The molecular orbital coefficients can be arranged as a matrix $\mathbf{C}$ which leads us to 
the Roothaan equation \cite{roothaan1951new, lehtola2020overview}
\begin{equation}\label{eq:roothaan}
    \mathbf{F} (\mathbf{C}) \mathbf{C} = \mathbf{S} \mathbf{C} \mathbf{\epsilon},
\end{equation}
where $\mathbf{F}$ is known as the Fock matrix which is a function of the molecular orbital coefficients $\mathbf{C}$, $\mathbf{S}$ is the overlap matrix,
and $\mathbf{\epsilon}$ is a diagonal matrix containing the molecular orbital energies.
The elements of both the Fock matrix and the overlap matrix are found by integrating over
combinations of the atomic orbitals.  
For example, an element of the overlap matrix is evaluated as:
\begin{equation}
    S_{\mu \nu} = \int \phi^*_\mu(\mathbf{r}) \phi_\nu(\mathbf{r}) d\mathbf{r}.
\end{equation}
The Gaussian representation of atomic orbitals has proven to be particularly
convenient as there are closed-form analytic solutions for
the various integrals needed to evaluate Equation \ref{eq:roothaan}.
MESS provides a basic implementation of these closed-form expressions
for all the relevant integrals needed for a typical simulation \cite{taketa1966gaussian}.

Equation \ref{eq:roothaan} is a non-linear generalised eigenvalue problem since the Fock matrix $\mathbf{F}$ depends on the 
molecular orbital coefficients $\mathbf{C}$.  Standard implementations will iteratively solve 
the Roothaan equation until finding the electronic energies converge to a fixed-point.
This process is known as the self-consistent field (SCF) method
\cite{lehtola2020overview}.

MESS provides an SCF solver but the preferred implementation performs
direct minimisation of the total energy functional:
\begin{equation}
    \min_{\mathbf{C}} E(\mathbf{C}),
\end{equation}
subject to the orthonormality constraint of the molecular orbitals:
\begin{equation}\label{eq:orthonormal}
    \mathbf{C}^\text{T} \mathbf{S} \mathbf{C} = \mathbf{1}.
\end{equation}
There are many ways to approach solving this constrained optimisation problem.  
We provide one such implementation in the \pythoninline{minimise} function of MESS
that uses the BFGS solver from the Optimistix library \cite{optimistix2024}.
The orthogonality constraint is incorporated by using the QR decomposition method
as outlined in the DQC library \cite{kasim2022dqc}.
Further, we demonstrate using first-order gradient methods in Section \ref{sec:examples} based on the Adam optimiser \cite{kingma2014adam}.

\section{Software Architecture}\label{sec:software}

MESS is designed to facilitate a new generation of hybrid electronic structure and machine learning 
simulations by providing a common framework that is hardware agnostic and optimised for current and future hardware. 
This forward-looking perspective is encapsulated by the term \textit{modern} in MESS. 
Our goals extend beyond mere integration: we aim to make electronic structure methods more accessible, 
transparent, and, crucially for research progress, more \textit{hackable}.

To accelerate progress towards these ambitious goals, we have adopted several key requirements, 
inspired by factors that have driven recent advancements in machine learning across various domains:

\begin{itemize}
\item \textbf{Hardware Acceleration:} Efficient utilisation of modern hardware accelerators 
is essential for successfully scaling computationally intensive simulations.
\item \textbf{Automatic Differentiation:} This capability enables seamless gradient calculations, which are vital for optimising complex models and exploring potential energy surfaces.
\item \textbf{High-Level Interpreted Language:}  A high-level language (e.g. Python) allows for rapid prototyping and experimentation, while modern machine learning compilers can mitigate the traditional performance concerns associated with interpreted languages.
\end{itemize}

Researchers accustomed to electronic structure codes implemented in compiled languages such as FORTRAN or C
might be skeptical about the performance of interpreted languages. 
However, the advent of machine learning compilers has significantly narrowed the gap between high-level code
and optimised machine code. 
These compilers excel at optimising tensor operations, the fundamental building blocks of many scientific computing 
and machine learning workloads, often achieving performance comparable to hand-tuned implementations
\cite{tillet2019triton}.


MESS is implemented entirely within the JAX framework, leveraging its rich ecosystem of libraries tailored
for both machine learning and scientific computing 
\cite{jax2018github, frostig2018compiling}. 
To encapsulate simulation parameters and provide an intuitive interface, we adopt the "callable dataclass"
abstraction from the Equinox library \cite{kidger2021equinox}. 
This approach enhances code readability and maintainability while aligning with 
JAX's functional programming paradigm.

JAX provides the following higher-order function transformations that 
we use throughout MESS:
\begin{itemize}
    \item \pythoninline{vmap}: Automatically vectorises functions, enabling efficient batch processing of multiple inputs.
    \item \pythoninline{grad}: Provides convenient and efficient Automatic Differentiation capabilities, allowing us to compute gradients of arbitrary order.
    \item \pythoninline{jit}: Just-in-time (JIT) compilation is a key feature of JAX that dynamically compiles Python functions into optimised machine code. By leveraging XLA, a high-performance compiler for linear algebra, \pythoninline{jit} significantly accelerates numerical computations in MESS.
    \item \pythoninline{pmap}: This transformation enables single-program, multiple-data (SPMD) parallel execution of functions across multiple devices (e.g. CPU, GPU, TPU, etc).
\end{itemize}

Broadly speaking, electronic structure simulation and machine learning model training share a common thread:  
both involve optimisation processes where the objective function that is being evaluated requires the interleaving
of linear algebra, reductions, and point-wise operations.  
This fundamental similarity suggests that software abstractions and optimisation techniques developed
for machine learning can be effectively applied to accelerate electronic structure calculations.

However, the rapidly evolving landscape of specialised hardware architectures and 
accompanying software frameworks presents a significant challenge for scientific software development. 
This constant change can dramatically influence the computational feasibility of various simulation methods. 
Choosing JAX as the foundation for MESS represents a strategic wager in this ``hardware lottery'' \cite{hooker2021hardware}. 
We posit that the substantial community interest in JAX-based machine learning models
will drive the continued development and optimisation of the framework.
Therefore, ensuring its optimisation on current hardware accelerators and further adaptation to future systems. 
This, in turn, will directly benefit MESS by enabling it to leverage cutting-edge hardware capabilities.

\section{Motivating Examples}\label{sec:examples}

The examples that follow have been adapted from executable notebooks that 
are provided as part of the documentation for MESS.
We have curated these examples to demonstrate 
how first-order gradient optimisation methods developed for machine learning
can be applied to tackle complex electronic structure simulations. 
In addition to this, a preliminary benchmark has shown a promising speedup of roughly 16x compared to performing the same energy minimisation with PySCF. 
While this initial result is encouraging, a more comprehensive benchmarking study is needed to 
definitively characterise the performance of MESS. 
We provide this benchmark as an executable notebook
\footnote{The reported 16X performance speedup was measured for a single water molecule with the PBE density functional and the 6-31G basis set.  MESS was executed on an NVIDIA A100 GPU accessed through Google Colab and PySCF was executed on the CPUs from the same runtime.  Full setup and measurements are available at \url{https://graphcore-research.github.io/mess/tour.html}}, inviting community feedback and ensuring the 
replicability of our findings.

\subsection{Direct Minimisation of Density Functional}\label{sec:direct-min}

A central problem of DFT simulations is finding the charge density $\rho(\mathbf{r})$
that minimises the total energy functional. 
This is a constrained optimisation since the charge density is derived from a 
basis of molecular orbitals that are required to be orthonormal to produce a valid solution.
Within MESS a single DFT calculation is setup by first defining the molecule, 
selecting a basis set, and building a \pythoninline{Hamiltonian} instance.
\begin{pythoncode}
  mol = from_pyquante("CH4")
  basis = basisset(mol, "6-31g")
  H = Hamiltonian(basis, xc_method="pbe")
\end{pythoncode}

This builds a single molecule of methane and represents the molecular orbitals 
with the 6-31G Pople basis set retrieved through the basis set exchange
\cite{pritchard2019new}.
The quantum-mechanical interactions are modelled within the Hamiltonian 
by selecting the \pythoninline{xc_method} argument which in the above example uses the 
widely PBE exchange-correlation functional \cite{perdew1996generalized}.
The PBE functional incorporates a gradient expansion of the electron density
and the implementation in MESS uses automatic differentiation to evaluate this gradient.  
In addition to various density functionals, MESS also implements the Hartree-Fock method, another widely used technique \cite{szabo2012modern}. 
Users can select Hartree-Fock by setting the \pythoninline{xc_method="hfx"} 
parameter to incorporate the Hartree-Fock exchange energy in the electron-electron interaction.

There are many possible approaches to solving the constrained optimisation for
the molecular orbital coefficients that minimise the total energy.  
The conjugate gradient method has been widely used in electronic structure simulations \cite{RevModPhys.64.1045}.
Here we take a different approach and minimise the electronic energy using the Adam implementation from the Optax optimisation library \cite{kingma2014adam, deepmind2020jax}.
The objective function we are minimising evaluates the total energy given 
a trial set of unconstrained parameters.  The total energy is comprised of an
electronic contribution that depends on the trial parameters and a static nuclear
contribution that represents the electrostatic energy of assembling the positively charged nuclear cores.
\begin{pythoncode}
@jax.jit
@jax.value_and_grad
def total_energy(Z):
    C = H.orthonormalise(Z)
    P = basis.density_matrix(C)
    return H(P) + nuclear_energy(mol)
\end{pythoncode}
Notice that the objective function explicitly applies the orthonormality constraint 
within the \pythoninline{orthonormalise} method of the Hamiltonian instance. 
The current implementation uses the QR decomposition algorithm as described in the 
DQC project to ensure that the orthonormality condition in Equation \ref{eq:orthonormal} is satisfied \cite{kasim2022dqc}.
Applying \pythoninline{value_and_grad} as a function decorator will transform the 
objective function to calculate the gradient with respect to the input at the 
same point as evaluating total energy.  
As described Section \ref{sec:software} the \pythoninline{jit} function decorator is applied to optimise
and compile the simulation so that it runs efficiently on the available hardware.

We use an arbitrary initial guess for the unconstrained parameters. Otherwise
the following energy minimisation will appear similar to a standard neural-network training loop
\begin{pythoncode}
Z = jnp.eye(basis.num_orbitals)
optimiser = optax.adam(learning_rate=0.1)
state = optimiser.init(Z)

for _ in range(200):
    e, grads = total_energy(Z)
    updates, state = optimiser.update(grads, state)
    Z = optax.apply_updates(Z, updates)
\end{pythoncode}

\begin{figure}[!ht]
    \centering
    \includegraphics[width=0.80\textwidth]{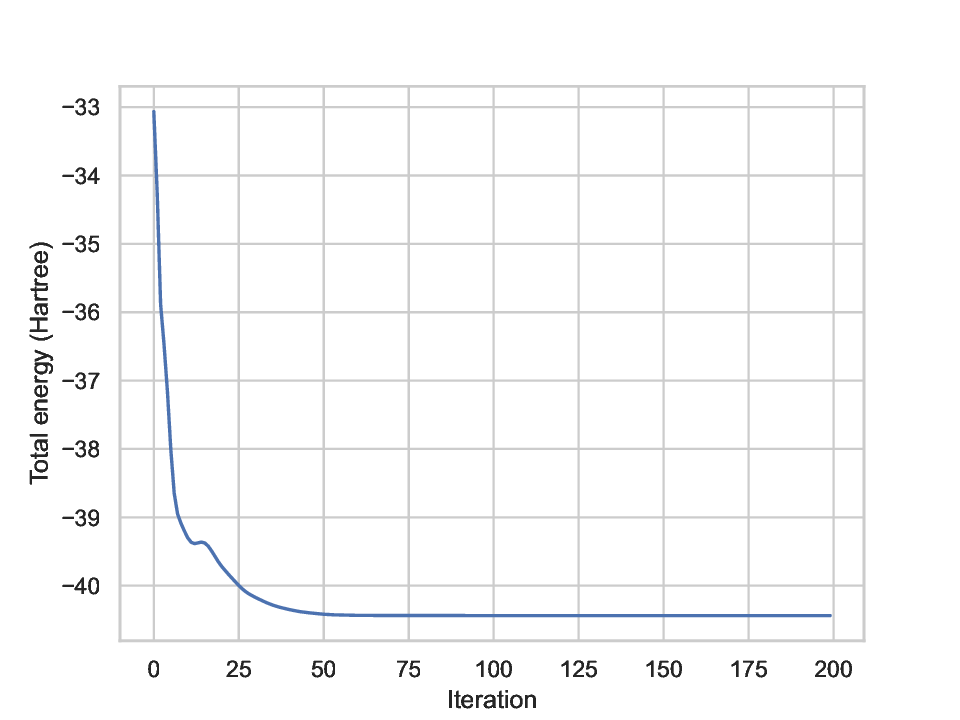}
    \caption{Total energy convergence for methane molecule with 6-31G basis set and PBE functional by direct optimisation using the Adam optimiser.}
    \label{fig:optim-loss}
\end{figure}

Figure \ref{fig:optim-loss} visualises how the energy varies through the optimisation process and shows the convergence of the total energy, which is exactly analogous 
to the analysis of the training loss curve in a neural-network.

\subsection{Batching over Conformations}

Training neural networks over batches of data is an essential component when efficiently
utilising massively parallel hardware accelerators. We can recast a standard electronic
minimisation problem as a batched one by using the JAX vectorising map (\pythoninline{vmap}).
By doing this we can parallelise the electronic structure optimisation over multiple
conformations of the same molecule.  Just like in training neural networks, this will
allow for more efficient utilisation of the accelerator which in turn allows
parallel sampling of potential energy surfaces using quantum-mechanical simulations.

We demonstrate this idea by calculating the molecular hydrogen dissociation curve using
a batch of hydrogen molecules where the bond length (H-H distance) is varied.  To setup
the batch we build a \pythoninline{Hamiltonian} for each bond length and stack the built modules to
create a batched-\pythoninline{Hamiltonian}.  This example uses the minimal STO-3G basis set and the simple
local density approximation (LDA) of DFT \cite{kohn1965self, parr1989density}. However, this batching formulation is not
limited to these choices for how the \pythoninline{Hamiltonian} is represented.
\begin{pythoncode}
def h2_hamiltonian(r: float):
    mol = Structure(
        atomic_number=np.array([1, 1]),
        position=np.array([[0.0, 0.0, 0.0], [r, 0.0, 0.0]]),
    )
    basis = basisset(mol, basis_name="sto-3g")
    return Hamiltonian(basis, xc_method="lda")


num_confs = 64
rs = np.linspace(0.6, 10, num_confs)
H = [h2_hamiltonian(r) for r in rs]
H = jax.tree.map(lambda *xs: jnp.stack(xs), *H)    
\end{pythoncode}

The following \pythoninline{energy} function evaluates the electronic energy of a single \pythoninline{Hamiltonian} 
for an unconstrained trial matrix $\mathbf{Z}$.  
The \pythoninline{vmap} function transformation converts this function that 
evaluates the energy for a single \pythoninline{Hamiltonian} to process a batched-\pythoninline{Hamiltonian}.
We also apply the \pythoninline{jit} function transformation to compile this function.
\begin{pythoncode}
@jax.jit
@jax.vmap
def energy(Z, H):
    C = H.orthonormalise(Z)
    P = H.basis.density_matrix(C)
    return H(P)    
\end{pythoncode}
As described in Section \ref{sec:direct-min}, we convert the constrained optimisation over molecular orbital coefficients into
an unconstrained one through the \pythoninline{orthonormalise} method of the Hamiltonian instance.
We initialise the optimisation to simultaneously minimise the energy of the batch of molecular conformations.
\begin{pythoncode}
num_orbitals = H[0].basis.num_orbitals
Z = jnp.tile(jnp.eye(num_orbitals), (num_confs, 1, 1))
optimiser = optax.adam(learning_rate=0.1)
state = optimiser.init(Z)
\end{pythoncode}

We define a straightforward loss function that simply takes the sum over the energy for each 
molecular conformation in the batch.
In basic terms, the optimisation will follow the gradient to minimise the loss. 
We use the transformation \pythoninline{value_and_grad}
as a function decorator on this loss function to evaluate the loss and the
corresponding gradient with automatic differentiation.
\begin{pythoncode}
@jax.value_and_grad
def loss_fn(z, h):
    return jnp.sum(energy(z, h))
\end{pythoncode}

Just as in Section \ref{sec:direct-min}, the optimisation looks familiar to a basic neural-network
training loop
\begin{pythoncode}
for _ in range(128):
    loss, grads = loss_fn(Z, H)
    updates, state = optimiser.update(grads, state)
    Z = optax.apply_updates(Z, updates)
\end{pythoncode}
Our loss function is only minimising the electronic energy. In order to calculate the total energy we
must add the electrostatic contribution that arises from each arrangement of nuclear charges.
We can evaluate this additive constant using the \pythoninline{vmap} transformation 
to calculate the nuclear energy for the entire batch of Hamiltonians
\begin{pythoncode}
E_nuclear = jax.vmap(nuclear_energy)(H.basis.structure)
E_total = energy(Z, H) + E_nuclear
\end{pythoncode}

\begin{figure*}
    \centering
    \includegraphics[width=1.0\textwidth]{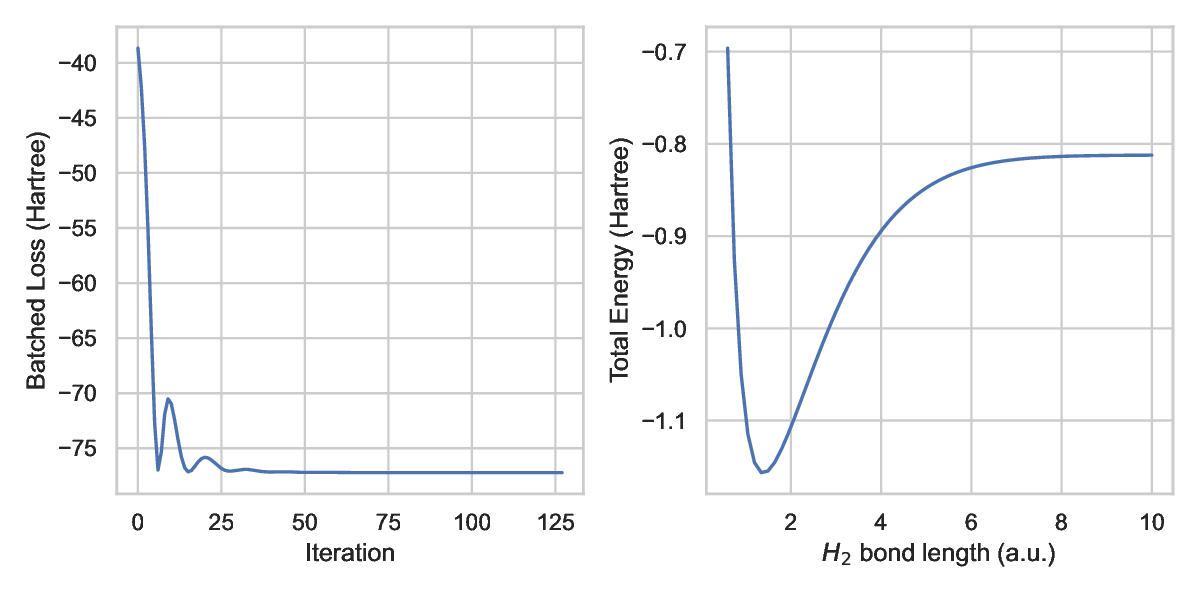}
    \caption{Batched evaluation of DFT energy for $H_2$ molecule within LDA approximation and the minimal STO-3G basis set.}
    \label{fig:batching}
\end{figure*}

In Figure \ref{fig:batching} we visualise the batched total loss as well as the 
converged bond dissociation curve which is evaluated entirely in parallel 
within our batch processing implementation for DFT. As far as we are aware, MESS is the first demonstration of a batched  implementation for DFT with full parallel evaluation.
We envision that batched implementations such as this can be used as a building block
for high-throughput DFT simulations for generating large-scale datasets.


\section{Concluding Thoughts}

We have presented MESS, a unified simulation environment designed to remove any friction for
adopting machine learning within electronic structure simulations.  
One of MESS's key features is the ability to leverage the JAX ecosystem, 
enabling hardware acceleration, automatic differentiation and optimisation through 
high-level function transformations.

Another promising direction for MESS is accelerating algorithmic research in electronic structure.
Significant progress has been made in reformulating DFT
to scale linearly with the number of atoms in the simulation rather than the cubic 
scaling of standard DFT implementations 
\cite{bowler2012methods, kohn1996density, prodan2005nearsightedness}.
These efforts have demonstrated that scaling electronic structure can be essential to elucidating 
complex phenomena across biochemistry and materials \cite{cole2016applications}.
We see MESS as an ideal environment for exploring novel electron density representations that 
can satisfy the requirements for achieving linear scaling DFT.


\printbibliography







\end{document}